\documentclass[runningheads]{llncs}
\pdfoutput=1
\usepackage[hyphens]{url}
\usepackage{graphicx}
\usepackage{color}
\usepackage{xcolor}
\usepackage[normalem]{ulem} 
\usepackage{amsmath,amssymb,amsfonts}
\usepackage{todonotes}
\graphicspath{{./figures/}}

\newcommand{\degree}{$^\circ$}
\newcommand{\safe}{\it safe}
\newcommand{\unsafe}{\it unrecov}
\newcommand{\delete}[1]{}

\newcommand{\myparagraph}[1]{\vspace{-1.5ex}\paragraph{#1}}

\begin{document}
\title{Neural Simplex Architecture}
\author{Dung T. Phan\inst{1} \and
Radu Grosu\inst{2} \and
Nils Jansen\inst{3} \and
Nicola Paoletti\inst{4} \and
Scott A. Smolka\inst{1} \and
Scott D. Stoller\inst{1}}
\authorrunning{D. Phan et al.}
\institute{Department of Computer Science, Stony Brook University, USA \and
Department of Computer Engineering, Technische Universit\"at Wien, Austria \and
Department of Software Science, Radboud University, Nijmegen, The Netherlands \and
Department of Computer Science, Royal Holloway, University of London, UK}
\maketitle              
\begin{abstract}
We present the \emph{Neural Simplex Architecture} (NSA), a new approach to runtime assurance that provides safety guarantees for neural controllers (obtained e.g.\ using reinforcement learning) of autonomous and other complex systems without unduly sacrificing performance.  NSA is inspired by the Simplex control architecture of Sha et al., but with some significant differences.  In the traditional approach, the advanced controller (AC) is treated as a black box; when the decision module switches control to the baseline controller (BC), the BC remains in control forever.  
There is relatively little work on switching control back to the AC, and there are no techniques for correcting the AC's behavior after it generates a potentially unsafe control input that causes a failover to the BC.   Our NSA addresses both of these limitations.  NSA not only provides safety assurances in the presence of a possibly unsafe neural controller, but can also improve the safety of such a controller in an online setting via retraining, without overly degrading its performance.
To demonstrate NSA's benefits, we have conducted several significant case studies in the continuous control domain.  These include a target-seeking ground rover navigating an obstacle field, and a neural controller for an artificial pancreas system.
\end{abstract}
\section{Introduction}\label{sec:intro}

Deep neural networks (DNNs) in combination with \emph{reinforcement learning} (RL) are increasingly being used to train powerful \emph{AI agents}.  Such agents have achieved unprecedented success in strategy games, including defeating the world champion in Go~\cite{silver2017masteringgo} and surpassing state-of-the-art chess and shogi engines~\cite{silver2017masteringchess}.
For these agents, safety is not an issue:
when a game-playing agent makes a mistake, the worst-case scenario is losing a game.  The same cannot be said for AI agents that control autonomous and other complex systems.  A mistake by an AI controller may cause physical damage to the controlled system and its environment, including humans.

In this paper, we present the \emph{Neural Simplex Architecture} (NSA), a new approach to runtime assurance that provides safety guarantees for AI controllers, including neural controllers, of autonomous and other complex systems without unduly sacrificing performance.  
NSA is inspired by Sha et al.'s Simplex control architecture~\cite{sha01using,seto98simplex},
where a pre-certified \emph{decision module} (DM) switches control from a high-performance but unverified (hence potentially unsafe) {\em advanced controller} (AC) to a verified-safe \emph{baseline controller} (BC)
if the AC produces an \emph{unrecoverable action}; i.e., an action that would lead the system within one time step to a state from which the BC is not guaranteed to preserve safety.

In the traditional Simplex approach, the AC is treated as a black box, and after the DM switches control to the BC, the BC remains in control forever. 
There is, however, relatively little work on 
switching control back to the AC \cite{johnson2016realtime,desai2019soter,vivekanandan2016simplex}, and there are no techniques to correct the AC after it generates an unrecoverable
control input. 

\begin{figure}[t]
\centering
\includegraphics[width=0.6\columnwidth]{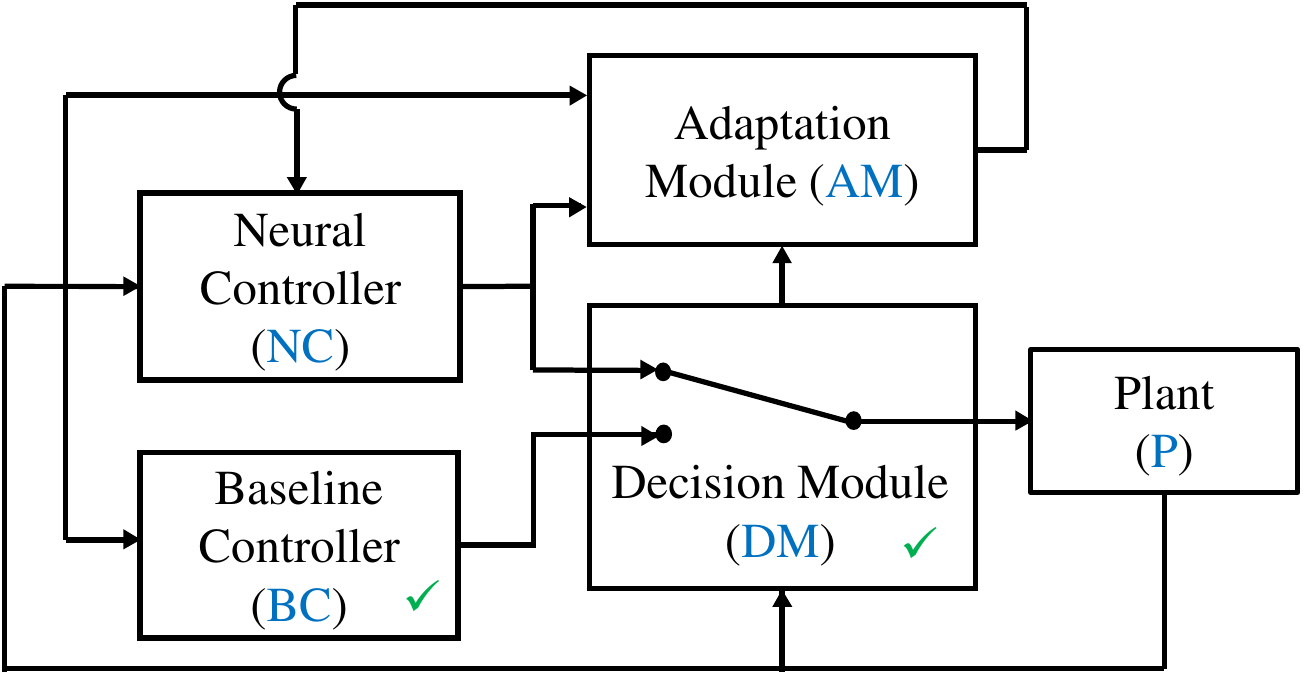}
\caption{The Neural Simplex Architecture. The green check marks indicate pre-certified components.}
\vspace{-0.5cm}
\label{fig:nsa}
\end{figure}

NSA, illustrated in Fig.~\ref{fig:nsa}, addresses both of these limitations.  
The high-performance \emph{Neural Controller} (NC) is a deep neural network (DNN) that given a plant state (or raw sensor readings),
produces a control input for the plant.  
NSA's use of an NC, as opposed to the black-box AC found in traditional Simplex,
allows for online retraining of the NC's DNN.  
Such retraining is performed by NSA's \emph{Adaptation Module} (AM) using RL techniques.  
For systems with large state spaces, it may be difficult to achieve thorough coverage during initial training of the NC.  
Online retraining has the advantage of focusing the learning on areas of the state space that are relevant to the actual system behavior, i.e., regions of the state space the system actually visits.

The AM seeks to eliminate unrecoverable actions from the NC's behavior, without unduly degrading its performance, and in some cases actually improving its performance.  While the BC is in control of the plant, the NC runs in shadow mode and is actively retrained by the AM. The DM can subsequently switch control back to the NC with high confidence that it will not repeat the same mistakes, permitting the mission to continue under the auspices of the high-performance NC.  Note that because NSA preserves the basic principles of Simplex architecture, it guarantees that the safety of the plant is never violated.


NSA addresses the problem of \emph{safe reinforcement learning}
(SRL)
\cite{garcia2015comprehensive,xiang2018verification}.
In particular, when the learning agent (the NC) produces an unrecoverable
action, the AM 
uses that action as a training sample (but does not execute it), with a large negative reward.
A comparison with related approaches to SRL is provided in Section~\ref{sec:related}.

We conducted an extensive evaluation of NSA on several significant example systems, including a target-seeking rover navigating through an obstacle field,
and a neural controller for an artificial pancreas. 
Our results on these case studies conclusively demonstrate NSA's benefits.

In summary, the main contributions of this paper are the following:
\vspace{-1.5ex}
\begin{itemize}
    \item We introduce the Neural Simplex Architecture, a new approach to runtime assurance that provides safety guarantees for neural controllers. 
\item  We address two limitations of the traditional Simplex approach, namely lack of established guidelines for switching control back to the AC so that mission completion can be attained; and lack of techniques for correcting the AC's behavior after a failover to the BC, so that reverse switching makes sense in the first place. 
\item\hspace{0.1em}  We provide a key insight into safe reinforcement learning (by demonstrating the utility of potentially unsafe training samples, when appropriately and significantly penalized),
along with a thorough evaluation of the NSA approach on two significant case studies.
\end{itemize}


\section{Background}\label{sec:bg}
\paragraph{Simplex Architecture.}
The main components of the Simplex architecture (AC, BC, DM) were introduced above.
%
The BC is certified to guarantee the safety of the plant only if it takes over control while the plant's state is within a \emph{recoverable region} $\mathcal{R}_{BC}$.  For example, consider the BC for a ground rover that simply applies maximum deceleration $a_{max}$.
The braking distance to stop the rover from a velocity $v$ is therefore $d_{br}(v) = v^2/(2\cdot a_{max})$.  The BC can be certified to prevent the rover from colliding with an obstacle if it takes over control in a state where $d_{br}(v)$ is less than the minimum distance $d_{min}$ to any obstacle.  The set of such states is the recoverable region of this BC.

A control input
is called \emph{recoverable} if it keeps the plant inside $\mathcal{R}_{BC}$ within the next time step.  Otherwise, the control input is called \emph{unrecoverable}.  The DM switches control to the BC when the AC produces an unrecoverable control input.  The DM's \emph{switching condition} determines whether a control input is unrecoverable.  We also refer to it as the {\em forward switching condition} (FSC) to distinguish it from the condition for \emph{reverse switching}, a new feature of NSA.

Techniques to determine the FSC include: (i)~shrink $\mathcal{R}_{BC}$ by an amount equal to a time step times the maximum gradient of the state with respect to the control input; then classify any control input as unrecoverable if the current state is outside this smaller region; (ii)~simulate a model of the plant for one time step if the model is deterministic and check whether the plant strays from $\mathcal{R}_{BC}$; (iii)~compute a set of states reachable within one time step and determine whether the reachable set contains states outside $\mathcal{R}_{BC}$.



\paragraph{Reinforcement Learning.} 
Reinforcement learning~\cite{sutton1998reinforcement} deals with the problem of how an \emph{agent} learns which sequence of \emph{actions} to take in a given \emph{environment} such that a cumulative \emph{reward} is maximized. At each time step $t$, the agent receives observation $s_t$ (the environment state) and reward $r_t$ from the environment and takes action $a_t$.  The environment receives action $a_t$ and emits observation $s_{t+1}$ and reward $r_{t+1}$ in response.  In the control of autonomous systems, the agent represents the controller, the environment represents the plant, and the state and action spaces are typically continuous. 

The goal of RL is to learn a \emph{policy} $\pi(a \,|\, s)$, i.e., a way of choosing an action $a$ having observed $s$, that maximizes the expected \emph{return} from the initial state, where the return at time $t$ is defined as the discounted sum of future rewards from $t$ (following policy $\pi$): $R_t = \sum_{k = t}^\infty \gamma^{k-t} r_{k+1}$; here $\gamma \in [0, 1]$ is a discount factor. For this purpose, RL algorithms typically involve estimating the action-value function $Q^\pi(s, a) = \mathbb{E} [R_t \mid s_t = s, a]$, i.e., the expected return for selecting action $a$ in state $s$ and then always following policy $\pi$; and the state-value function $V^\pi(s) = \mathbb{E}[R_t \mid s_t = s]$, i.e., the expected return starting from $s$ and following $\pi$.

While early RL algorithms were designed for discrete state and action spaces, recent \emph{deep RL} algorithms, such as TRPO~\cite{schulman2015trust}, DDPG~\cite{lillicrap2015continuous}, A3C~\cite{mnih2016asynchronous}, and ACER~\cite{wang2016sample}, have emerged as promising solutions for RL-based control problems in continuous domains. These algorithms leverage the expressiveness of deep neural networks (DNNs) to represent policies and value functions. 

\section{Neural Simplex Architecture}\label{sec:nsa}

In this section, we discuss the main components of NSA, namely the neural controller (NC), the adaptation module (AM), and the reverse switching logic.  These components in particular are not found in the Simplex control architecture.

The dynamics of the plant, i.e., the system under control, is given by 
$s_{t+1}=f(s_t,a_t)$,
where $s_t \in \mathcal{S}$ is the state of the plant at time $t$, $\mathcal{S}\subseteq \mathbb{R}^n$ is the real-valued state space, $f$ is a possibly nonlinear function, and
$a_t \in \mathcal{A}$ is the
control input to the plant
at time $t$,
with $\mathcal{A}\subseteq\mathbb{R}^m$ the
action space. This equation specifies a deterministic dynamics, even though our approach equally supports nondeterministic ($s_{t+1}\in f_{nd}(s_t,a_t)$) 
and stochastic ($s_{t+1}\sim f_{st}(s \mid s_t,a_t$)) 
plant dynamics.  
We assume full observability, i.e., that the BC and NC have access to the full state of the system $s_t$.\footnote{In case of partial observability, the full state can typically be reconstructed from sequences of past states and actions, but this process is error-prone.}

We denote with $\mathrm{DM}_t\in\{\mathrm{NC}, \mathrm{BC}\}$ the state of the decision module at time $t$:  $\mathrm{DM}_t=\mathrm{NC}$ ($\mathrm{DM}_t=\mathrm{BC}$) indicates that the neural (baseline) controller is in control. Let $a_t^{\rm NC}$ and $a_t^{\rm BC}$ denote the action computed by the NC and the BC, respectively. The final action $a_t$ performed by the NSA agent depends on the DM state: $a_t=a_t^{\rm NC}$ if $\mathrm{DM}_t=\mathrm{NC}$, $a_t=a_t^{\rm BC}$ if $\mathrm{DM}_t=\mathrm{BC}$. 

Let $\beta$ be the BC's control law, i.e., $a_t^{\rm BC} = \beta(s_t)$.  For a set of unsafe states $\mathcal{U}\subseteq \mathcal{S}$, the \emph{recoverable region} is 
the largest set $\mathcal{R}_{BC}$ such that $s\in \mathcal{R}_{BC} \Rightarrow f(s,\beta(s)) \in \mathcal{R}_{BC}$ and $\mathcal{R}_{BC}\cap \mathcal{U}=\emptyset$. For $s\in \mathcal{S}$, $a\in \mathcal{A}$, the forward switching condition must satisfy $f(s,a)\not\in \mathcal{R}_{BC} \Rightarrow \text{FSC}(s,a)$.

\myparagraph{The Neural Controller.}
The NC is represented by a DNN-based policy $\pi_{\theta_t}$, where $\theta_t$ are the current DNN parameters. The policy maps the current state into a proposed action $a_t^{\rm NC} = \pi_{\theta_t}(s_t)$. We stress the time dependency of the parameters because adaptation and retraining of the policy is a key feature of NSA.  As for the dynamics $f$, our approach supports stochastic policies ($a_t^{\rm NC} \sim \pi(a \mid s_t,\theta_t)$). 


The NC can be obtained using any RL algorithm.  We used DDPG with the safe learning strategy of penalizing unrecoverable actions, as discussed in Section~\ref{sec:safe-rl}.  DDPG is attractive as it 
works with deterministic policies, and allows uncorrelated samples to be added to the pool of samples for training or retraining.  The latter property is important because it allows us to collect disconnected samples of what the NC would do while the plant is under the BC's control, and to use these samples for online retraining of the NC.

\myparagraph{Adaptation and Retraining.}

The AM is used to retrain the NC in an online manner while the BC is in control of the plant (due to NC-to-BC failover).  The main purpose of this retraining is to make the NC less likely to trigger the FSC, thereby allowing it to remain in control for longer periods of time, thereby improving overall system performance.

Techniques that we consider for online retraining of the NC include supervised learning and reinforcement learning.  In supervised learning, state-action pairs of the form $(s, a)$ are required for training purposes.  The training algorithm uses these examples to teach the NC safe behavior.  The control inputs produced by the BC can be used as training samples, although this will train the NC to imitate BC's behavior, which may lead to a loss in performance.

We therefore prefer SRL for online retraining, with a reward function that penalizes unrecoverable actions and rewards recoverable, high-performing ones.  
The reward function for retraining can be designed as follows.

\begin{equation}
r(s, a, s') = \begin{cases}
     r_{\unsafe},& \text{if } \text{FSC}(s, a)\\
    r_{{\it perf}}(s, a, s'),              & \text{otherwise}
\end{cases}
\end{equation}
where 
$r_{{\it perf}}(s, a, s')$ is a performance-related reward function, and $r_{\unsafe}$ is a negative number used to penalize unrecoverable actions. The benefits of this approach to SRL are discussed in Section~\ref{sec:safe-rl}. 

The AM retrains the NC at each time step the BC is in control by maintaining a pool of retraining samples of the form $(s_t,a_t^{\rm NC},s',r')$, where $a_t^{\rm NC}$ is the NC-proposed action, $s' = f(s_t, a_t^{\rm NC})$ is the state that the system would evolve to if the NC was in control, and $r'=r(s,a_t^{\rm NC},s')$ is the corresponding reward. I.e., samples are obtained by running the NC in shadow mode: when BC is in control, the AM obtains a retraining sample by running a simulation of the system for one time step and applying $a_t^{\rm NC}$, while the actual system evolves according to the BC action $a_t^{\rm BC}$. 

The AM updates the NC's parameters $\theta_t$ as follows: 
\[
\theta_t = 
\begin{cases}
\mathrm{RL}(\theta_{t-1},(s_t,a_t^{\rm NC},s',r')), & \text{if $\mathrm{DM}_{t} = \mathrm{BC}$}\\
\theta_{t-1}, & \text{otherwise}
\end{cases}
\]
where $\mathrm{RL}$ is the deep RL algorithm chosen for NC adaptation. 
Note that as soon as the DM switches control to the BC after the NC has produced an unrecoverable action (see also the Switching logic paragraph below), a corresponding retraining sample for the NC's action is added to the pool. 


We evaluated a number of variants of this procedure by making different choices along the following dimensions.
\begin{enumerate}
    \item Start retraining with an empty pool of samples or with the pool created during the initial training of the NC.
    \item Add (or do not add) exploration noise to NC's action when collecting a sample. With exploration noise, the resulting action is $a_t^{\mathrm{NC}} + \nu_t$, where $\nu_t$ is a random noise term. 
    Note that we consider noise only when NC is running in shadow mode (BC in control), as directly using noisy actions to control the plant would degrade performance.
    \item Collect retraining samples only while BC is in control 
    or at every time step.  In both cases, the action in each training sample is the action output by NC (or a noisy version of it); we never use BC's action in a training sample.  Also, in both cases, the retraining algorithm for updating the NC 
    is run only while the BC is in control.
\end{enumerate}
We found that reusing the pool of training samples (DDPG's so-called experience replay buffer) from initial training of the NC helps 
evolve the policy in a more stable way,
as retraining samples gradually replace  initial training samples in the sample pool.  Another benefit of reusing the initial training pool is that the NC can be immediately retrained without having to wait for enough samples to be collected online.
We found that adding exploration noise to NC's actions in retraining samples, and collecting retraining samples at every time step, both increase the benefit of retraining. This is because these two strategies provide more diverse samples and thereby help achieve more thorough exploration of the state-action space.


\myparagraph{Switching logic.}

%
\begin{figure}[t]
    \centering
    \includegraphics[scale=0.45]{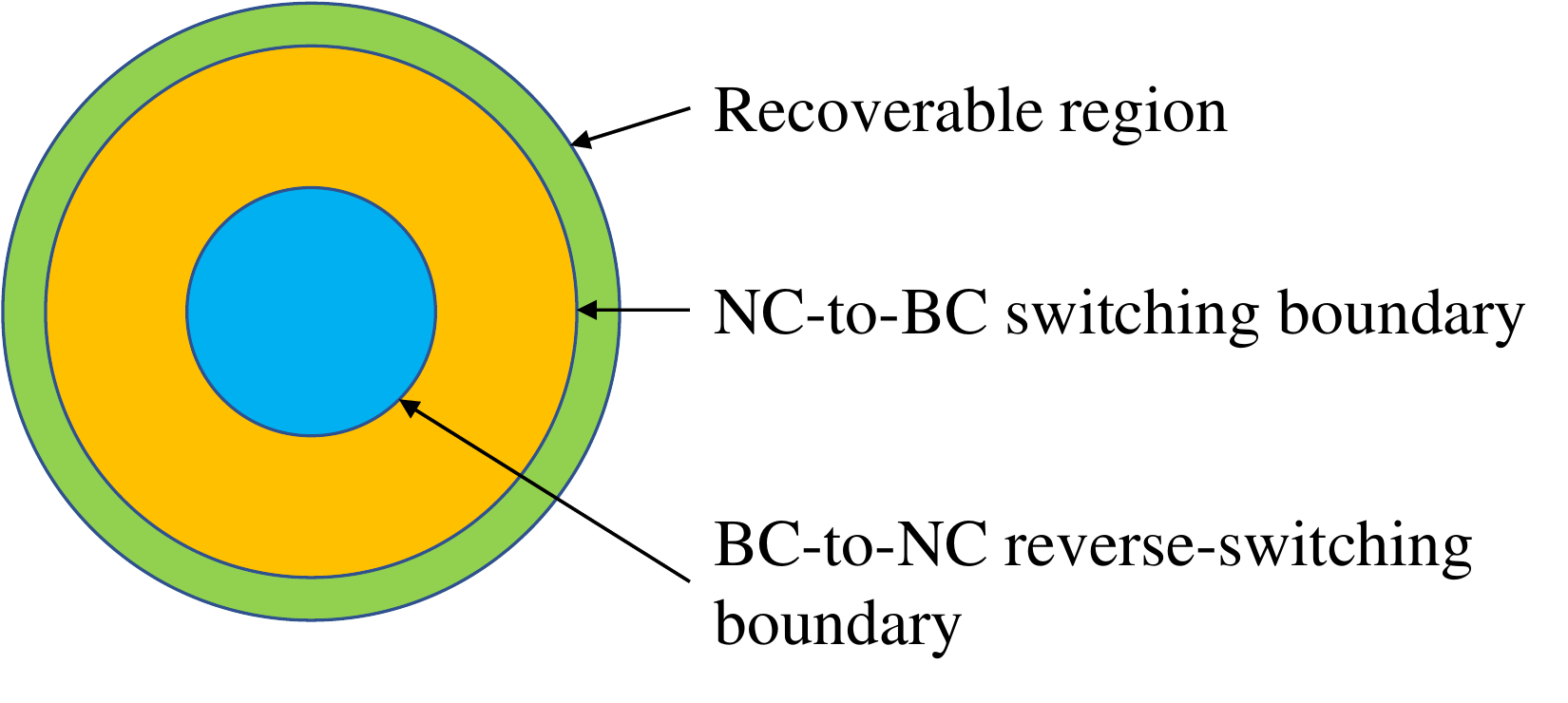}
    \caption{Switching boundaries. The blue region is a subset of the orange area, which in turn is a subset of the green region.}
    \vspace*{-.5cm}
    \label{fig:switching-boundaries}
\end{figure}

NSA includes \emph{reverse switching} from the BC to the retrained NC.
An additional benefit of well-designed reverse switching is that it lessens the burden on the BC to achieve performance objectives, leading to a simpler BC design that focuses mainly on safety. 
Control of the plant is returned to the NC when the reverse switching condition (RSC) is true in the current state. 
We can summarize NSA's switching logic by describing the evolution of the DM state $\mathrm{DM}_t$. NSA begins with the NC in control, i.e., $\mathrm{DM}_t=\mathrm{NC}$ for $t\leq 0$. For $t>0$, the DM state is given by:
\[
\mathrm{DM}_t = 
\begin{cases}
\mathrm{BC}, & \text{if $\mathrm{DM}_{t-1}=\mathrm{NC}$ and FSC$(s_t,a_t^{\rm NC})$}\\
\mathrm{NC}, & \text{if $\mathrm{DM}_{t-1}=\mathrm{BC}$ and RSC$(s_t)$}\\
\mathrm{DM}_{t-1}, & \text{otherwise}
\end{cases}
\]
To ensure safety when returning control to the NC, the FSC must not hold if the RSC is satisfied, i.e.,  
$\text{RSC}(s) \Rightarrow \neg \text{FSC}(s,a)$, for all $s\in\mathcal{S}, a\in\mathcal{A}$. 

We seek to develop reverse switching logic that returns control to NC when it is safe to do so and which avoids frequent back-and-forth switching between the BC and NC.
We propose two such approaches.  
One is to reverse-switch if a forward switch will not occur in the near future. This can be checked by simulating the composition of the NC and plant for $T$ time steps, and reverse-switching if the FSC does not hold within this time horizon.\footnote{For nondeterministic (stochastic) systems, a
(probabilistic) model checker can be used instead of a simulator, but this approach may be computationally expensive.}  Formally, $\text{RSC}(s_t) = \bigwedge_{t'=t}^{t+T} \neg \text{FSC}(s'_{t'},\pi_{\theta_{t}}(s'_{t'}))$, where $s'_{t}=s_t$ and $s'_{t'+1}=f(s'_{t'},\pi_{\theta_{t}}(s'_{t'}))$. 
This approach, used in our inverted pendulum and artificial pancreas case studies, prevents frequent switching.

A simpler approach is to reverse-switch if the current plant state is sufficiently far from the NC-to-BC switching boundary; see Fig.~\ref{fig:switching-boundaries}.  Formally, $\text{RSC}(s_t) = \sup\{d(s_t,s') \mid s' \in \mathcal{S}, \ \text{FSC}(s',\pi_{\theta_t}(s'))\} > \epsilon$, where $d$ is a metric on $\mathbb{R}^n$ and $\epsilon\in \mathbb{R}^{+}$ is the desired distance. This approach is used in our rover 
case study.

We emphasize that the choice of RSC does not affect safety and is application-dependent.   Note that both of our approaches construct an RSC that is stricter than a straight complement of the FSC.  This helps avoid excessive switching.
In our experiments,
we empirically observed that the system behavior was not very sensitive to the exact
value of $T$ or $\epsilon$; so choosing acceptable values for them is not difficult.


\section{Safe Reinforcement Learning with Penalized Unrecoverable Continuous Actions} \label{sec:safe-rl}

We evaluate the use of two policy-gradient algorithms for safe reinforcement learning in NSA. The first approach filters the learning agent's unrecoverable actions before they reach the plant.  For example, when the learning agent, i.e., the NC, produces an unrecoverable action, a runtime monitor~\cite{fulton2018safe} or a preemptive shield~\cite{alshiekh2018safe} replaces it with a recoverable one to continue the trajectory.  The recoverable action is also passed to the RL algorithm to update the agent and training continues with the rest of the trajectory.

In the second approach, when the learning agent produces an unrecoverable action,
we assign a penalty (negative reward) to the action, use it as a training sample, and then use recoverable actions to safely terminate the trajectory (but not to train the agent).  Safely terminating the trajectory is important in cases where for example the live system is used for training.  We call this approach \emph{safe reinforcement learning with penalized unrecoverable continuous actions} (SRL-PUA).  By ``continuous'' here we mean real-valued action spaces, as in~\cite{dalal2018safe}.  Other SRL approaches such as~\cite{alshiekh2017safearxiv} use discrete actions.

To compare the two approaches, we used the DDPG and TRPO algorithms to train neural controllers for an inverted pendulum (IP) control system. 
Details about our IP case study, including the reward function and the BC used to generate recoverable actions, can be found in the Appendix.
 
We used the implementations of DDPG and TRPO in rllab~\cite{duan2016benchmarking}.  For TRPO, we trained two DNNs, one for the mean and the other for the standard deviation of a Gaussian policy.  Both DNNs have two fully connected hidden layers of 32 neurons each and one output layer.  The hidden layers use the \texttt{tanh} activation function, and the output layer is linear.  For DDPG, we trained a DNN that computes the action directly from the state.  The DNN has two fully connected hidden layers of 32 neurons each and one output layer.  The hidden layers use the \texttt{ReLU} activation function, and the output layer uses \texttt{tanh}.  We followed the choice of activation functions in the examples accompanying rllab. 

For each algorithm, we ran two training experiments.  In one experiment, we reproduce the filtering approach; i.e., we replace an unrecoverable action produced by the learning agent with the BC's recoverable action, use the latter as the training sample, and continue the trajectory.  We call this training method SRL-BC. In the other experiment, we evaluate the SRL-PUA approach.  Note that both algorithms explore different trajectories by resetting the system to a random initial state whenever the current trajectory is terminated.  We set the maximum trajectory length to 500 time steps, meaning that a trajectory is terminated when it exceeds 500 time steps.

We trained the DDPG and TRPO agents on a total of one million time steps.  After training, we evaluated all trained policies on the same set of 1,000 random initial states.  During evaluation, if an agent produces an unrecoverable action, the trajectory is terminated.  
The results are shown in Table~\ref{tbl:ip_policy_comparison}.  For both algorithms, the policies trained with recoverable actions (the SRL-BC approach) produce unrecoverable actions in all test trajectories, while the SRL-PUA approach, where the policies are trained with penalties for unrecoverable actions, does not produce any such actions.  As such, the latter policies achieve superior returns and trajectory lengths (they are able to safely control the system the entire time).

In the above experiments, we replaced unrecoverable actions with actions generated by a deterministic BC, whereas the monitoring~\cite{fulton2018safe} and preemptive shielding~\cite{alshiekh2017safearxiv} approaches allow unrecoverable actions to be replaced with random recoverable ones, an approach we refer to as SRL-RND.  To show that our conclusions are independent of this difference, we ran one more experiment with each learning algorithm, in which we replaced each unrecoverable action with an action selected by randomly generating actions until a recoverable one is found.  The results, shown in Table~\ref{tbl:ip_rand_recoverable}, once again demonstrate that training with only recoverable actions is ineffective.
Compared to filtering-based approaches (SRL-BC in Table~\ref{tbl:ip_policy_comparison} and SRL-RND in Table~\ref{tbl:ip_rand_recoverable}), the SRL-PUA approach yields a 25- to 775-fold improvement in the average return.
\begin{table}
    \centering
    \begin{tabular}{@{\hspace{-1pt}}r@{\hspace{2pt}}|@{\hspace{2pt}}c@{\hspace{2pt}}|@{\hspace{2pt}}c@{\hspace{2pt}}|@{\hspace{2pt}}c@{\hspace{2pt}}|@{\hspace{2pt}}c@{\hspace{2pt}}|}
        \multicolumn{1}{c}{} & \multicolumn{2}{c}{\textbf{TRPO}} & \multicolumn{2}{c}{\textbf{DDPG}}\\
        & \textbf{SRL-BC} & \textbf{SRL-PUA} & \textbf{SRL-BC} & \textbf{SRL-PUA} \\
        \textbf{Unrec Trajs} & 1,000 & 0 & 1,000 & 0 \\
        \textbf{Comp Trajs} & 0 & 1,000 & 0 & 1,000 \\
        \textbf{Avg.\ Return} & 112.53 & 4,603.97 & 61.52 & 4,596.04 \\
        \textbf{Avg.\ Length} & 15.15 & 500 & 14.56 & 500 \\
    \end{tabular}
    \medskip
    \caption{Policy performance comparison. \textbf{SRL-BC}: policy trained with BC's actions replacing unrecoverable ones. \textbf{SRL-PUA}: policy trained with penalized unsafe actions. \textbf{Unrec Trajs}: \texttt{\#} trajectories terminated
    due to an unrecoverable action.  \textbf{Comp Trajs}: \texttt{\#} trajectories that reach the limit of 500 time steps. \textbf{Avg.\ Return} and \textbf{Avg.\ Length}: average return and trajectory length over 1,000
    trajectories.}
    \label{tbl:ip_policy_comparison}
\end{table}

\begin{table}
\vspace{-1cm}
    \centering
    \begin{tabular}{@{\hspace{-1pt}}r@{\hspace{2pt}}|@{\hspace{2pt}}c@{\hspace{2pt}}|@{\hspace{2pt}}c@{\hspace{2pt}}|@{\hspace{2pt}}c@{\hspace{2pt}}|@{\hspace{2pt}}c@{\hspace{2pt}}|}
        \multicolumn{1}{c}{} & \multicolumn{2}{c}{\textbf{TRPO}} & \multicolumn{2}{c}{\textbf{DDPG}}\\
        & \textbf{SRL-RND} & \textbf{SRL-PUA} & \textbf{SRL-RND} & \textbf{SRL-PUA} \\
        \textbf{Unrec Trajs} & 1,000 & 0 & 1,000 & 0 \\
        \textbf{Comp Trajs} & 0 & 1,000 & 0 & 1,000 \\
        \textbf{Avg. Return} & 183.36 & 4,603.97 & 5.93 & 4,596.04 \\
        \textbf{Avg. Length} & 1.93 & 500 & 14 & 500 \\
    \end{tabular}
    \medskip
    \caption{Policy performance comparison. \textbf{SRL-RND}: policy trained with random recoverable actions replacing unrecoverable ones.}
    \vspace{-1cm}
    \label{tbl:ip_rand_recoverable}
\end{table}

\section{Case Studies}
An additional case study, the Inverted Pendulum, along with further details about the case studies presented in this section can be found in the Appendix.

\subsection{Rover Navigation}\label{sec:cs-rover}


We consider the problem of navigating a rover to a predetermined target location while avoiding collisions with static obstacles.  The rover is a circular disk of radius $r$.  It has a maximum speed $v_{max}$ and a maximum acceleration $a_{max}$.  The maximum braking time is therefore $t_{br\_max} = v_{max} / a_{max}$, and the maximum braking distance is $d_{br\_max} = v_{max}^2/(2\cdot a_{max})$.  The control inputs are the accelerations $a_x$ and $a_y$ in the $x$ and $y$ directions, respectively.  The system uses discrete-time control with a time step of $dt$.

The rover
has $n$ distance sensors whose detection range is $l_{max}$.  The sensors are placed evenly around the perimeter of the rover; i.e., the center lines of sight of two adjacent sensors form an angle of $2\pi/n$.  The rover can only move forwards, so its orientation is the same as its heading angle.  The state vector for the rover is $[x, y, \theta, v, l_1, l_2, ..., l_n]$, where $(x, y)$ is the position, $\theta$ is the heading angle, $v$ is the velocity, and the $l_i$'s are the sensor readings.
%
\begin{figure}
    \centering
    \includegraphics[trim={0cm 0.2cm 0cm 0cm},clip]{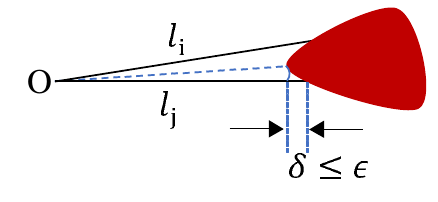}
    \caption{Illustration of assumptions about obstacle shapes.}
    \label{fig:obstacle_assumption}
    \vspace{-0.5cm}
\end{figure}

We assume the sensors have a small angular field-of-view so that each sensor reading reflects the distance from the rover to an obstacle along the sensor's center line of sight.  If a sensor does not detect an obstacle, its reading is $l_{max}$. 

We also assume that when the sensor readings of two adjacent sensors $s_i$ and $s_j$ are $l_i$ and $l_j$, respectively, then the (conservative) minimum distance to any obstacle point located in the cone formed by the center lines of sight of $s_i$ and $s_j$ is $\min\lbrace l_i, l_j\rbrace - \epsilon$. Here, $\epsilon$ is a constant that limits by how much an obstacle can protrude into the blind spot between $s_i$ and $s_j$'s lines of sight; see Fig.~\ref{fig:obstacle_assumption}. 


A state $s$ of the rover is \emph{recoverable} if starting from $s$, the baseline controller (BC) can brake to a stop at least distance $d_{\safe}$ from any obstacle.  Let the braking distance in state $s$ be $d_{br}(s) = v^2/(2\cdot a_{max})$, where $v$ is the rover's speed in $s$.  Then $s$ is recoverable if the minimum sensor reading $l_{min}$ in state $s$ is at least $d_{\safe} + d_{br}(s) + \epsilon$. 

The FSC holds when the control input $u_{NC}$ proposed by the NC will put the rover in an unrecoverable state in the next time step.  We check this condition
by simulating the rover for one time step with $u_{NC}$ as the control input, and by then determining if $l_{min} < d_{\safe} + d_{br}(s) + \epsilon$.

The RSC is $l_{min} \ge m\cdot v_{max} \cdot dt + d_{\safe} + d_{br\_max} + \epsilon$, ensuring that the FSC does not hold for the next $m-1$ time steps.
Parameter $m$ can be chosen to reduce excessive back-and-forth switching between the NC and BC.



The BC performs the following steps:
1)~Apply the maximum braking power $a_{max}$ until the rover stops.
2)~Randomly pick a safe heading angle $\theta$ based on the current position and sensor readings.
3)~Rotate the rover until its heading angle is $\theta$.
4)~Move with heading angle $\theta$ until either the FSC becomes true (this is checked after each time step by the BC itself), in which case the BC is re-started at Step~1, or the RSC becomes true (this is checked by the DM), in which case the NC takes over.


    
    



%
%
\myparagraph{Experimental Results.}
Parameter values used: $r = 0.1\,$m, $v_{max} = 0.8\,$m/s, $a_{max} = 1.6\,$m/$\text{s}^2$, $l_{max} = 2\,$m, $n = 32$, $d_{\safe} = 0.2\,$m, $\epsilon = 0.01\,$m, $m = 5$, $dt = 0.1\,$s. The target is a circular disk at location $(0, 0)$ with a radius of 0.1m.  The obstacle field, which is fixed during training and testing, consists of 12 circular obstacles with a minimum radius of 0.25m.
Rover initial position $(x_0, y_0)$ is randomized in the area $[-5, 5]\times[-5, 5]$.~\footnote{Although the obstacles are fixed, the NC still generalizes well (but not perfectly) to random obstacle fields not seen during training, as shown in this video https://youtu.be/ICT8D1uniIw.}
We assume that the sensor field-of-view is at least $7.25^\circ$, thereby satisfying the assumption that an obstacle does not protrude more than $\epsilon$ into the blind spot between adjacent sensors. See also  Fig.~\ref{fig:obstacle_assumption}.
The NC is a DNN with two \texttt{ReLU} hidden layers, each of size 64, and a \texttt{tanh} output layer.  We used the DDPG algorithm for both initial training and online retraining of the NC.  
For initial training, 
we ran DDPG for 5~million time steps. 
The reward function for initial training and online retraining is:
\newcommand{\dist}{\text{DT}}
\begin{equation}
r(s, a, s') = 
\begin{cases}
    -20,000, & \text{if } \text{FSC}(s, a)\\
     10,000, & \text{if } \dist(s) \le 0.2\\
    -1 - 20\cdot\dist(s), & \text{otherwise}
\end{cases}
\end{equation}
where $\text{FSC}(s, a)$ is the forward switching condition and $\dist(s)$ is the center-to-center distance from the rover to the target in state $s$.  The rover is considered to have reached the target if $\dist(s) \le 0.2$, as, recall, the target is a circular disk with radius of 0.1m and the radius $r$ of the rover is 0.1m.  If the action $a$ triggers the forward switching logic, it is penalized by assigning it a negative reward of -20,000.  If $a$ causes the rover to reach the target, it receives a positive reward of 10,000.  All other actions are penalized by an amount proportional to the distance to the target, encouraging the agent to reach the target quickly.
Our experiments with online retraining use the same DDPG settings as in initial training, except that we initialize the AM's pool of retraining samples with the pool created by initial training, instead of an empty pool.  The pool created by initial training contains one million samples; this is the maximum pool size, which is a parameter of the algorithm.  When creating retraining samples, the AM adds Gaussian noise to the NC's actions.  The NC's actions are collected (added to the pool) at every time step, regardless of which controller is in control; thus, the AM also collects samples of what the NC would do while the BC is in control.  

We ran the NSA instance starting from 10,000 random initial states.  Out of 10,000 trajectories, forward switching occurred in 456 of them.  Of these 456 trajectories, the BC was in control for a total of 70,974 time steps.  This means there were 70,974 ($\sim$71K) retraining updates to the NC.  To evaluate the benefits of online retraining, we compared the performance of 
the NC after initial training and after 20K, 50K, and 71K online updates.  We evaluated the performance of each of these controllers (by itself, without NSA) by running it from the same set of 1,000 random initial states.

The results in Table~\ref{tbl:rover_retrain_comparison} show that after 71K retraining updates, the NC outperforms the initially trained version on every metric.  Table~\ref{tbl:rover_retrain_comparison} also shows that the NC's performance increases with the number of retraining updates, thus demonstrating that NSA's online retraining not only improves the safety of the NC, but also its performance.

\begin{table}[t]
    \centering
    
    \begin{tabular}{r|c|c|c|c|c}
         & \textbf{IT} & \textbf{20K RT} & \textbf{50K RT} & \textbf{71K RT} \\
         \textbf{FSCs} & 100 & 79 & 43 & 8 \\
         \textbf{Timeouts} & 35 & 49 & 50 & 22 \\
         \textbf{Targets} & 865 & 872 & 907 & 970 \\
         \textbf{Avg.\ Ret.} & -9,137.3 & -9,968.8 & -5,314.6 & -684.0 \\
         \textbf{Avg.\ Len.} & 138.67 & 142.29 & 156.13 & 146.56 \\
    \end{tabular}
    \medskip
    \caption{Benefits of online retraining ($\sim$71K NC updates in total) for ground rover navigation. \textbf{IT}: results for initially trained NC.  \textbf{20K RT}, \textbf{50K RT}, \textbf{71K RT}: results for NC after 20K, 50K, 71K retraining updates. 
    All  controllers evaluated on same set of 1,000 random initial states.  \textbf{FSCs}: \texttt{\#} trajectories in which FSC becomes true.  \textbf{Timeouts}: \texttt{\#} trajectories that reach the
    limit of 500 time steps without reaching target or having FSC become true. \textbf{Targets}: \texttt{\#} trajectories that reach the target.  \textbf{Avg.\ Ret.} and \textbf{Avg.\ Len.}: average return and average trajectory length
    over all 1,000 trajectories.}
    \vspace{-0.75cm}
    \label{tbl:rover_retrain_comparison}
\end{table}


We resumed initial training to see if this would produce similar improvements.  Specifically, we continued the initial training for an additional 71K, 1M, and 3M samples.  The results, included in the Appendix, show that extending the initial training slowly improves both the safety and performance of the NC but requires substantially more updates.  71K retraining updates provide significantly more benefits than even 3M additional samples of initial training.

\subsection{Artificial Pancreas}\label{sec:cs-ap}

The artificial pancreas (AP) is used to control blood glucose (BG) levels in Type~1 diabetes patients through automated delivery of insulin. We use the linear plant model of~\cite{chen2019} to describe the physiological state of the patient. The main state variable of interest is $G$, which is the difference between the reference BG ($7.8$ mmol/L) and the patient's BG. The control action, i.e., the insulin input, is denoted by $u$.
Further details of this model, including its ODE dynamics, can be found in the Appendix.

The AP should maintain BG levels within the safe range of~4 to~11 mmol/L.  In particular, it should avoid hypoglycemia (i.e., BG levels below the safe range), which can lead to severe health consequences. Hypoglycemia occurs when the controller overshoots the insulin dose. Insulin control is uniquely challenging because the controller cannot take a corrective action to counteract an excessive dose; its most extreme safety measure is to turn off the insulin pump. Hence, the baseline controller for the AP sets $u=0$. 



We intentionally under-train the initial NC so that it exhibits low performance and produces unrecoverable actions. Low-performing AP controllers may arise in practice for several reasons, e.g., when the training-time model parameters do not match the current real-life patient parameters. 

The reward function $r$ is designed to penalize deviations from the reference BG level, as captured by state variable $G$. We assign a positive reward when $G$ is close to zero (within $\pm 1$), and we penalize larger deviations with a 5$\times$ factor for mild hyperglycemia ($1 < G \leq 3.2$), a 7$\times$ factor for mild hypoglycemia ($-3.8 \leq G < -1$), 9$\times$ for strong hyperglycemia ($G > 3.2$), and 20$\times$ for strong hypoglycemia ($G < -3.8$). 
The other constants are chosen to avoid jump discontinuities in the reward function.
\begin{equation*}
r(s, u, s') = \begin{cases}
     10 - |G'|, & \text{if } |G'|\leq 1 \\ 
     14 - 5\cdot|G'|, & \text{if }  1 < G' \leq 3.2\\ 
     26.8 - 9\cdot|G'|, & \text{if }  G' > 3.2\\
     16 - 7\cdot|G'|, & \text{if }  -3.8 \leq G' < -1\\
     65.4 - 20\cdot|G'| &  \text{otherwise}
\end{cases}
\end{equation*}
where $G'$ is the value of $G$ in state $s'$. 



An AP plant state $s$ is \emph{recoverable} if under the control of the BC, a state where $G' < -3.8$ cannot be reached starting from $s$. This condition can be checked by simulation. 
The FSC holds when the NC's action leads to an unrecoverable state in the next time step. For reverse switching, we
return
control to the NC if the FSC does not hold within time $T=10$ from the current state. 

\myparagraph{Experimental Results.}
To produce an under-trained NC, we used 107,000 time steps of initial training.  
We ran NSA on the under-trained controller on 10,000 trajectories, each starting from a random initial state.  Among the first 400 trajectories, 250 led to forward switching and hence retraining. The retraining was very effective, as forward switching did not occur after the first 400 trajectories.

As in the other case studies we conducted, we then evaluated the benefits of retraining by comparing the performance of the initially trained NC and the retrained NC 
on trajectories starting from the same set of 1,000 random initial states.  The results are given in Table~\ref{tbl:ap_retrain_comparison}.
Retraining greatly improves the safety of the NC: the initially trained controller reaches an unrecoverable state in all 1,000 of these trajectories, while the retrained controller never
does.
The retrained controller's performance is also significantly enhanced, with an average return 2.9 times that of the initial controller. 

\begin{table}
    \centering


    \begin{tabular}{r|c|c|}
         & \textbf{Initially Trained} & \textbf{Retrained} \\
         \textbf{Unrecov Trajs} & 1,000 & 0 \\
         \textbf{Complete Trajs} & 0 & 1,000\\
         \textbf{Avg.\ Return} & 824 & 2,402\\
         \textbf{Avg.\ Length} & 217 & 500\\
    \end{tabular}
    \medskip
    \caption{Benefits of retraining for the AP case study. There were 61 updates to the NC. Row labels are as per Table~\ref{tbl:ip_policy_comparison}.}
    \vspace{-1cm}
    \label{tbl:ap_retrain_comparison}
\end{table}

\section{Related Work}\label{sec:related}

The original Simplex architecture did not consider
reverse switching. In~\cite{seto99acase,seto98simplex}, when the AC produces an unrecoverable action, it is disabled until it is manually re-enabled.  
It is briefly mentioned in~\cite{johnson2016realtime} that reverse switching should be performed only when the FSC is false, and that a stricter RSC might be needed to prevent frequent switching, but the paper does not pursue this idea further.  
A more general approach to reverse switching, which uses reachability analysis to determine if the plant is safe in the next two time steps irrespective of the controller, is presented in~\cite{desai2019soter}.  This approach results in more conservative reverse switching conditions, as it does not take the behavior of the AC into account, unlike one of the approaches we propose.   The idea of reverse switching when the AC's outputs are stabilized is briefly mentioned in~\cite{vivekanandan2016simplex}.

Regarding approaches to safe reinforcement learning (SRL), we refer the reader to two recent comprehensive literature reviews \cite{garcia2015comprehensive,xiang2018verification}.
Bootstrapping of policies that are known to be safe in certain environments is employed in~\cite{DBLP:conf/aaai/SimaoS19}, 
while~\cite{garcia2019probabilistic} restricts exploration to a portion of the state space close to an optimal, pre-computed policy.
%
%

In~\cite{alshiekh2018safe}, the authors synthesize a \emph{shield} (a.k.a.\ \emph{post-posed shield}) from a temporal-logic safety specification based on knowledge of
the system dynamics.  The shield monitors and corrects an agent's actions to ensure safety.  This approach targets systems with finite state and action spaces.
Suitable finite-state abstractions are needed for infinite-state systems.
In~\cite{DBLP:journals/corr/abs-1904-07189}, the shield-based approach is extended to stochastic systems.
In contrast, NSA's policy-gradient-based approach is directly applicable to
systems with infinite state spaces and continuous action spaces.
%
%
%
%

In~\cite{fulton2018safe}, the authors use formally verified runtime monitors
in the RL training phase to constrain the actions taken by the learning agent to a set of safe actions.  
The idea of using the learned policy together with a known-safe fallback policy in the deployed system is mentioned, but further details are not provided. In contrast, we discuss in detail how the NSA approach guarantees runtime safety and how SRL is is used for online retraining of the NC.  In~\cite{fulton2019verifiably}, a verification-preserving procedure is proposed for learning updates to the environment model when SRL is used and the exact model is not initially known.
The approach to SRL is mainly taken from~\cite{fulton2018safe}, so again the learned policy is not guaranteed safe. Note that the SRL approach of~\cite{fulton2018safe,fulton2019verifiably} allows the training algorithm to speculate when the plant model is deviating from reality.

Other approaches to SRL incorporate formal methods to constrain the SRL exploration process.
These include the use of (probabilistic) temporal logic~\cite{DBLP:journals/corr/WenET15,hasanbeig2018logically,DBLP:conf/icaart/MasonCKB17}, ergodicity-based notions of safety~\cite{moldovan2012safe}, and providing probably approximately correct (PAC) guarantees~\cite{Fu-RSS-14}.  All of these techniques work on finite state spaces.

In~\cite{chow2018lyapunov}, the authors
use Lyapunov functions in the framework of constrained Markov decision processes
to guarantee policy safety during training.  They
focus on policy-iteration and Q-learning
for discrete state and action problems.  Their approach is currently not applicable to policy-gradient algorithms, such as the DDPG algorithm used in our experiments, nor continuous state/action problems.
Lyapuanov functions are also used in~\cite{berkenkamp2017safe} for SRL, but it likewise cannot be used for
policy-gradient algorithms.

In~\cite{tessler2018reward}, the authors propose Reward Constrained Policy Optimization (RCPO), where a per-state weighted penalty term
is added to the reward function.  Such weights are
updated during training.  RCPO is shown to almost surely converge to a
solution, but does not address the problem of  guaranteeing safety during training.
In contrast, we penalize unrecoverable actions and safely terminate the current trajectory to ensure plant safety.  

In~\cite{achiam2017constrained}, the authors present the Constrained Policy Optimization
algorithm for constrained MDPs, which guarantees safe exploration during training.
CPO only ensures approximate satisfaction of constraints and provides an upper bound on the cost associated with constraint violations.
In~\cite{ohnishi2018safety}, the authors use control barrier functions (CBFs) for SRL. 
Whenever the learning agent produces an unsafe action, it is minimally perturbed to preserve safety. In contrast, in NSA, when the NC proposes an unsafe action, the BC takes over and the NC is retrained by the AM.
CBFs are also used in~\cite{cheng2019end}.

Similar to the shield-based method, a safety layer is inserted between the policy and the plant in~\cite{dalal2018safe}.  Like the CBF approach, the safety layer uses quadratic programming to minimally perturb the action to ensure safety.  There are, however, no formal guarantees of safety because of the data-driven linearization of the constraint function.

\section{Conclusions}\label{sec:conclusions}

We have presented the Neural Simplex Architecture for assuring the runtime safety of systems with
neural controllers.  NSA features an adaptation module that retrains the NC in an online fashion, seeking to eliminate its faulty behavior without unduly sacrificing performance.
NSA's reverse switching capability allows control of the plant to be returned to the NC after a failover to BC, thereby allowing NC's performance benefits to come back into play.  We have demonstrated the utility of NSA on three significant case studies in the continuous control domain. 

As future work, we plan to investigate methods for establishing statistical bounds on the degree of improvement that online retraining yields in terms of safety and performance of the NC.  
We also plan to incorporate techniques from the L1Simplex architecture~\cite{wang2013l1simplex} to deal with deviations of the plant model's behavior from the actual behavior.

\paragraph{\bf Acknowledgments.}

We thank the anonymous reviewers for their helpful comments.
This material is based upon work supported in part by %
NSF grants %
  CCF-191822, 
  CPS-1446832, 
  IIS-1447549, 
  CNS-1445770, 
  and CCF-1414078, 
  FWF-NFN RiSE Award, 
    and ONR grant N00014-15-1-2208. 
Any opinions, findings, and conclusions or recommendations expressed in
this material are those of the author(s) and do not necessarily reflect
the views of these organizations.

\bibliographystyle{splncs04}
\bibliography{references}
\newpage
\appendix

\section*{Appendix}
\section{Inverted Pendulum Case Study}\label{sec:ip}



\begin{figure}
	\centering
	\includegraphics[width=0.3\columnwidth]{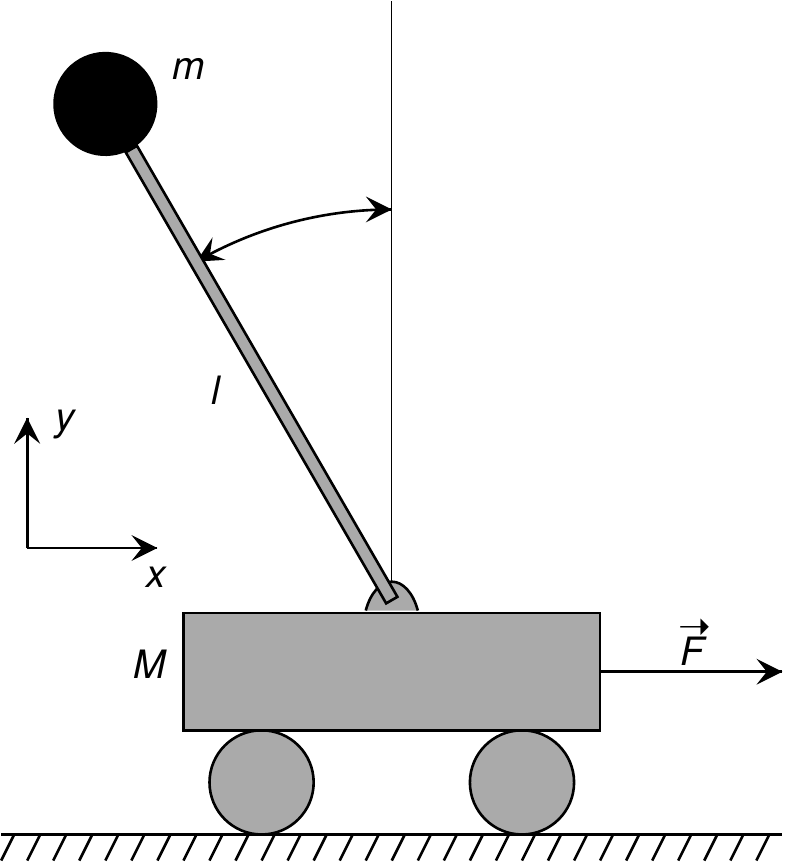}
    \caption{Schematic of the inverted pendulum on a cart. Source: Wikipedia.}
    \label{fig:ip-schematic}
\end{figure}

We consider the classic control problem of keeping an inverted pendulum upright on a movable cart.  We describe the problem briefly here; a detailed exposition is available  in~\cite{seto99acase}.  
A schematic diagram appears in Fig.~\ref{fig:ip-schematic}.  
The linearized dynamics is given by $\dot{\mathrm{x}} = \mathrm{A}\cdot \mathrm{x} + \mathrm{B} \cdot V_a$, 
where $\mathrm{x} = \left[p, v, \theta, \omega\right]^T$ is the state vector consisting of the cart position $p$, cart velocity $v$, pendulum angle $\theta$, and pendulum angular velocity $\omega$, and control input $V_a$ is the armature voltage applied to the cart's motor.  
The constant matrix $\mathrm{A}$ and constant vector $\mathrm{B}$ used in~\cite{seto99acase,johnson2016realtime} and our case study are

\begin{equation*}
\mathrm{A} = \begin{bmatrix}
    0 & 1 & 0 & 0 \\
    0 & -10.95 & -2.75 & 0.0043 \\
    0 & 0 & 0 & 1 \\
    0 & 24.92 & 28.58 & -0.044 \\
\end{bmatrix}
\quad 
\mathrm{B} = \begin{bmatrix}0 \\ 1.94 \\ 0 \\ -4.44\end{bmatrix}
\end{equation*}

The safety constraints for this system are $p \in [-1, 1]$ m, $v \in [-1, 1]$ m/s, and $\theta \in [-15, 15]$\degree.  The control input $V_a$ is constrained to be in $[-4.95, 4.95]$ V.  Although $\omega$ is unconstrained, its physical limits are implicitly imposed by the constraints on $v$.  The control objective is to keep the pendulum in the upright position, i.e., $\theta = 0$.


The BC is a linear state feedback controller of the form $V_a = \mathrm{K}\cdot \mathrm{x}$, with the objective of stabilizing the system to the setpoint $\mathrm{x} = 0$.  We obtain this controller by using the linear matrix inequality (LMI) approach described in~\cite{seto99acase}. 
The LMI approach computes a vector $\mathrm{K}$ and a matrix $\mathrm{P}$ such that:
\begin{itemize}
\item When the system state is inside the ellipsoid $\mathrm{x}^T\mathrm{P}\mathrm{x} \le 1$, all safety constraints are satisfied.
\item When the system starts in a state inside the ellipsoid $\mathrm{x}^T\mathrm{P}\mathrm{x} \le 1$ and uses BC's control law $V_a = \mathrm{K}\cdot \mathrm{x}$, it will remain in this ellipsoid forever.
\end{itemize}

The gain vector $\mathrm{K}$ and matrix $\mathrm{P}$ produced by the LMI approach for the described inverted pendulum system are:

\begin{equation*}
\mathrm{K} = \begin{bmatrix}
0.4072 \\ 7.2373 \\ 18.6269 \\ 3.6725
\end{bmatrix}
\mathrm{P} = \begin{bmatrix}
1.0520 & 0.2580 & 1.2082 & 0.1988 \\
0.2580 & 2.2108 & 4.6631 & 1.0090 \\
1.2082 & 4.6631 & 33.9334 & 4.0269 \\
0.1988 & 1.0090 & 4.0269 & 0.8424 
\end{bmatrix}
\end{equation*}

The matrix $\mathrm{P}$ defines a \emph{recoverable} region $\mathcal{R}_{BC} = \lbrace \mathrm{x} \mid \mathrm{x}^T\mathrm{P}\mathrm{x} \le 1 \rbrace$.  The forward switching condition is that the control input $V_a$ will drive the system outside $\mathcal{R}$ in the next time step.
For the RSC, the DM simulates the NC for 10 time steps starting from the current state, and switches to the NC if there are no safety violations within this time horizon. 


The inverted pendulum problem can be considered ``solved'' by many reinforcement learning algorithms, due to its small state-action space.  To demonstrate the online retraining capability of NSA's adaptation module, we intentionally under-train a neural controller, so that it produces unrecoverable actions.  We used the DDPG algorithm with the following reward function, where $v'$ and $\theta'$ are the velocity and pendulum angle in state $\mathrm{x}'$:

\begin{equation*}
r(\mathrm{x}, V_{a}, \mathrm{x}') = \begin{cases}
     0,              & \text{if } \text{FSC}(x, V_{a}) \\ 
     10 - 10\, v'^2 - (1 - \cos{\theta'}), &  \text{otherwise}
\end{cases}
\end{equation*}

This reward function encourages the controller to (i)~keep the pendulum upright via the penalty term $-(1 - \cos{\theta'})$, and (ii)~minimize the movement of the cart via the penalty term $10\cdot v'^2$.  The total distance travelled by the cart is one performance metric where the NC is expected to do better than the BC.  Whenever the forward switching condition becomes true, the execution terminates.  Therefore, the neural controller should also learn to respect (not trigger) the FSC, in order to maximize the discounted cumulative reward.  Each execution is limited to 500 time steps.






\paragraph{Experimental Results.}
We under-trained an NC by training it for only 500,000 time steps.  The DNN for the NC has the same architecture as the DDPG DNN.
For our retraining experiments, we created an NSA instance consisting of this NC and the BC described above.  We reused the initial training pool that has 500,000 samples, added Gaussian noise to NC's actions in retraining samples, and collected retraining samples at every time step.

We ran the NSA instance starting from 2,000 random initial states.  Out of the 2,000 trajectories, forward switching occurred in 28 of them,
resulting in
the BC being in control for a total of 4,477 time steps and therefore 4,477 retraining updates to the NC.  Notably, there was only one forward switch in the last 1,000 trajectories; this shows that the retraining during the first 1,000 trajectories significantly improved the NC's safety. 

To evaluate the overall benefits of retraining, we ran the initially trained NC and the retrained NC starting from the same set of 1,000 random initial states.  The results, given in Table~\ref{tbl:ip_it_rt}, show that after just 4,477 retraining updates, the retrained NC completely stops producing unrecoverable actions,
which significantly improves the average return, increasing it by a factor of~2.7.

\begin{table}
    \centering
    \begin{tabular}{r|c|c|}
         & \textbf{Initially Trained} & \textbf{Retrained} \\
         \textbf{Unrecov Trajs} & 976 & 0 \\
         \textbf{Complete Trajs} & 24 & 1,000 \\ 
         \textbf{Avg.\ Return} & 1,711.17 & 4,547.11 \\
         \textbf{Avg.\ Length} & 203.26 & 500
    \end{tabular}
    \medskip
    \caption{Benefits of retraining for the inverted pendulum, based on 1,000 trajectories used for evaluation.  Unrecov Trajs: \texttt{\#} trajectories terminated because of an unrecoverable action.  Complete Trajs: \texttt{\#} trajectories that reach limit of 500 time steps. Avg.\ Return and Avg.\ Length: average return and average trajectory length  over all 1,000 trajectories.}
    \vspace{-1cm}
    \label{tbl:ip_it_rt}
\end{table}



\section{Rover Navigation Case Study}
\begin{figure}
\centering
\includegraphics[width=0.6\columnwidth,trim={4cm 3.9cm 4cm 1.5cm},clip,width=0.75\columnwidth]{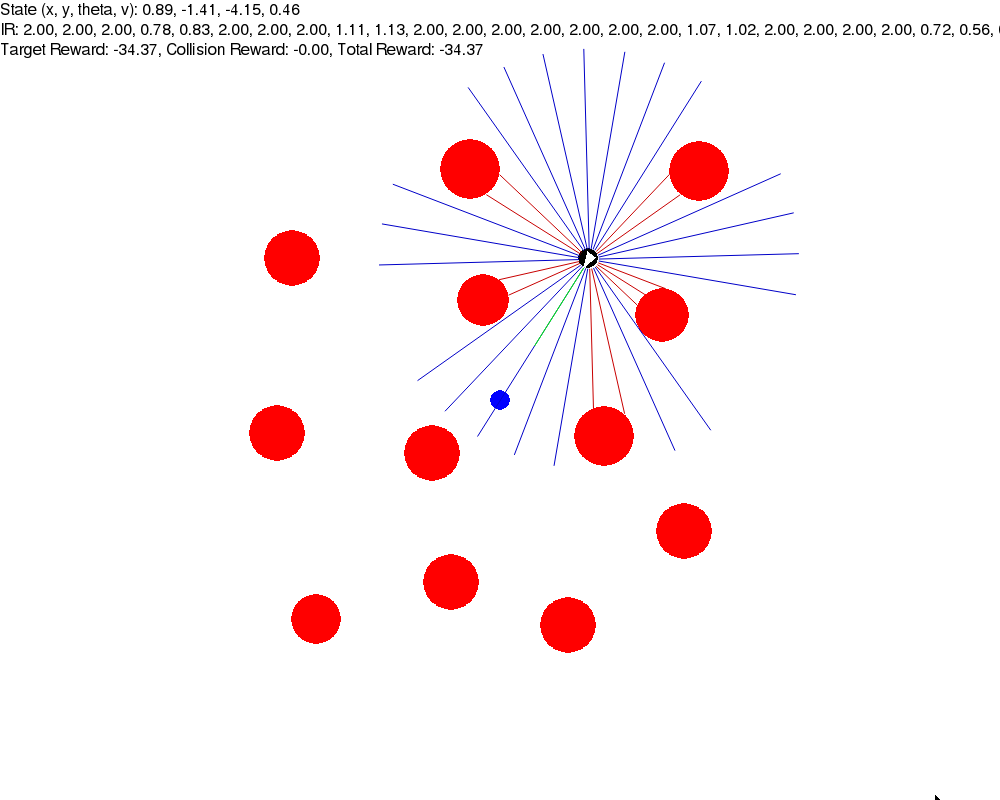}
\caption{Training and testing setup for the rover case study. The red disks are obstacles, the black dot with an inscribed white triangle is the rover, and the blue dot is the target.  The spokes coming out of the rover represent the distance sensors. The rover's heading angle is shown by the orientation of the inscribed triangle.  The length of the green line, which is a prefix of the spoke for the sensor pointing directly forward, is proportional to the rover's speed. }
\label{fig:rover_setup}
\end{figure}


Figure~\ref{fig:rover_setup} shows the training and testing setup for the rover case study.  A video showing how the initially trained NC navigates the rover through the same obstacle field used in training is available at \url{https://youtu.be/nTeQ4eHF-fQ}.  The video shows that the NC is able to reach the target most of the times.  However, it occasionally drives the rover into unrecoverable states.  Note that all supplementary videos are sped up 10x.  If we pair this NC with the BC in an NSA instance, the rover never enters unrecoverable states.  A video showing this NSA instance in action with reverse switching enabled and online retraining disabled is available at \url{https://youtu.be/XbJKrnuxcuM}.  Note that in this video, we curated only interesting trajectories where switches occurred.  In the video, the rover is black when the NC is in control; it turns green when the BC is in control.

The initially trained NC also performs reasonably well on random obstacle fields not seen during training.  A video of this is available at \url{https://youtu.be/ICT8D1uniIw}.  The rover under NC control is able to reach the target most of the time.  However, it sometimes overshoots the target, suggesting that we may need to vary the target position during training.  We plan to investigate this as future work.

The results of extending the initial training appear in Table~\ref{tbl:rover_extended_it}.  Extending the initial training slowly improves both the safety and the performance of the NC but requires substantially more updates.  Comparing with Table~3 in the paper shows that 71K of retraining updates in NSA provide significantly more benefits than even 3M additional updates of initial training.  NSA's retraining is much more effective, because it samples more unrecoverable actions while the plant is under BC's control, and because it tends to focus retraining on regions of the state-action space of greatest interest, especially regions near the forward switching boundary and regions near the current state.  In contrast, trajectories in initial training start from random initial states.

\begin{table}
    \centering
    \begin{tabular}{r|c|c|c|c|c}
         & \textbf{IT} & \textbf{71K EIT} & \textbf{1M EIT} & \textbf{3M EIT} \\
         \textbf{FSCs} & 100 & 108 & 108 & 78 \\
         \textbf{Timeouts} & 35 & 224 & 78 & 43 \\
         \textbf{Targets} & 865 & 668 & 814 & 879 \\
         \textbf{Avg.\ Ret.} & -9,137.3 & -12,448.3 & -9,484.8 & -3,320.4 \\
         \textbf{Avg.\ Len.} & 138.67 & 215.7 & 137.75 & 124.26 \\
    \end{tabular}
    \medskip
    \caption{Extended initial training performance. 71K EIT, 1M EIT, and 3M EIT: results for NC after 71K, 1M, and 3M updates during extended initial training.  All controllers evaluated using same set of 1,000 random initial states used for 
    Table~3 in the paper.}
    \vspace{-1cm}
    \label{tbl:rover_extended_it}
\end{table}

\section{Artificial Pancreas Case Study}\label{app:cs-ap}


The artificial pancreas (AP) is a system for controlling blood glucose (BG) levels in Type 1 Diabetes patients through the automated delivery of insulin. Here we consider the problem of controlling the {\em basal insulin}, i.e., the insulin required in between meals. We consider a deterministic linear model (adapted from~\cite{chen2015towards}) to describe the physiological state of the patient. The dynamics are given by:
\begin{align}
\dot{G}(t) = & -p_1\cdot G(t) -p_2 \cdot I(t) + p_3 \label{eq:bmodel1}\\
\dot{I}(t) = & -k_e \cdot I(t) + \frac{k_a}{V_I}x(t)\label{eq:bmodel4}\\
\dot{x}(t) = & -k_a \cdot x(t) + u(t)\label{eq:bmodel5}
\end{align}
where $G(t)$ is the difference between the reference BG, $r_{BG}=7.8$ mmol/L, and the patient's BG; $u(t)$ (mU/min) is the insulin input (i.e., the control input); $x(t)$ (mU) is the insulin mass in the subcutaneous compartment; and $I(t)$ is the plasma insulin concentration (mU/L). Parameters $p_1,\ldots,p_3, k_e,k_a,V_I >0$ are patient-specific.  

The AP should keep BG levels within safe ranges, typically 4 to 11 mmol/L, and in particular it should avoid hypoglycemia (i.e., BG levels below the safe range), a condition that leads to severe health consequences. Hypoglycemia happens when the controller overshoots the insulin dose. What makes insulin control uniquely challenging is the fact that the controller cannot take a corrective action to counteract an excessive dose; its most drastic safety measure is to shut off the insulin pump. For this reason, the baseline controller for the AP sets $u=0$.

For this case study, we assume that the controller can observe the full state of the system, and thus, the corresponding policy is a map of the form $(G,I,x)\mapsto u$. We perform discrete-time simulations of the ODE system with a time step of $1$. 


Similarly to the inverted pendulum problem, we intentionally under-train the NC so that it produces unrecoverable actions.  This results in an AP controller with poor performance.  Controllers with poor performance may arise in practice for a variety of reasons, including the (common) situation where the physiological parameters used during training poorly reflect the patient's physiology. 

The reward function is designed to penalize deviations from the reference BG level. Such a deviation is promptly given by the state variable $G$. We give a positive reward when $G$ is close to zero (within $\pm 1$), and we penalize larger deviations with a 5$\times$ factor for mild hyperglycemia ($1 < G \leq 3.2$), a 7$\times$ factor for mild hypoglycemia ($-3.8 \leq G < -1$), 9$\times$ for strong hyperglycemia ($G > 3.2$), and 20$\times$ for strong hypoglycemia ($G < -3.8$). The other constants are chosen to avoid jump discontinuities in the reward function.
\begin{equation}
r(s, u, s') = \begin{cases}
     10 - |G'|, & \text{if } |G'|\leq 1 \\ 
     14 - 5\cdot|G'|, & \text{if }  1 < G' \leq 3.2\\ 
     26.8 - 9\cdot|G'|, & \text{if }  G' > 3.2\\
     16 - 7\cdot|G'|, & \text{if }  -3.8 \leq G' < -1\\
     65.4 - 20\cdot|G'| &  \text{otherwise}
\end{cases}
\end{equation}
where $G'$ is the value of $G$ in state $s'$. This reward function is inspired by the asymmetric objective functions used in previous work on model predictive control for the AP~\cite{gondhalekar2016periodic,cmsb2017}.


A state $s$ is \emph{recoverable} if under the control of the BC ($u=0$), the system does not undergo hypoglycemia ($G' < -3.8$) in any future state starting from $s$. This condition is checked by simulating the system from $s$ with $u=0$ until $G$ starts to increase: as one can see from the system dynamics (\ref{eq:bmodel1}--\ref{eq:bmodel5}), this is the point at which $G$ reaches its minimum value under the BC.

The FSC holds when the control input proposed by the NC leads to an unrecoverable state in the next time step. For reverse switching, we use the default strategy of returning control to the NC if applying the NC for a bounded time horizon $T=10$ from the current state does not produce a state satisfying the FSC.

\paragraph{Experimental Results}
To produce an under-trained NC, we used 107,000 time steps of initial training.  For retraining, we used the same settings as in the inverted pendulum case study. 

\begin{table}
    \centering


    \begin{tabular}{r|c|c|}
         & \textbf{Initially Trained} & \textbf{Retrained} \\
         \textbf{Unrecov Trajs} & 1,000 & 0 \\
         \textbf{Complete Trajs} & 0 & 1,000\\
         \textbf{Avg.\ Return} & 824 & 2,402\\
         \textbf{Avg.\ Length} & 217 & 500\\
    \end{tabular}\medskip
    \caption{Benefits of retraining for the AP case study. There were 61 updates to the NC. Row and column labels are as per Table~\ref{tbl:ip_it_rt}.}
    \vspace{-1cm}
    \label{app:tbl:ap_retrain_comparison}
\end{table}

We ran the NSA instance for 10,000 trajectories.  Among the first 400 trajectories, 250 led to forward switches and hence retraining. The retraining that occurred in those 250 trajectories was very effective, because forward switching never occurred after the first 400 trajectories.  As we did for the other case studies, we then evaluated the benefits of retraining by comparing the performance of the initially trained NC and the retrained NC (by themselves, without NSA) on trajectories starting from the same set of 1,000 random initial states.  The results are in Table~\ref{app:tbl:ap_retrain_comparison}.  We observe that retraining greatly improves the safety of the NC: the initially trained controller reaches an unrecoverable state in all 1,000 of these trajectories, while the retrained controller never reaches an unrecoverable state. The retrained controller's performance is also significantly enhanced, with an average return 2.9 times higher than that of the initial controller.

\end{document}